# User Experience of a Smart Factory Robot: Assembly Line Workers Demand Adaptive Robots

Astrid Weiss[1] and Andreas Huber[1]

**Abstract.** This paper reports a case study on the User Experience (UX) of an industrial robotic prototype in the context of human-robot cooperation in an automotive assembly line. The goal was to find out what kinds of suggestions the assembly line workers – who actually use the new robotic system – propose in order to improve the human-robot interaction (HRI). The operators working with the robotic prototype were interviewed three weeks after the deployment using established UX narrative interview guidelines. Our results show that the cooperation with a robot that executes predefined working steps actually impedes the user in terms of flexibility and individual speed. This results in a change of working routine for the operators, impacts the UX, and potentially leads to a decrease in productivity. We present the results of the interviews as well as first thoughts on technical solutions in order to enhance the adaptivity and subsequently the UX of the human-robot cooperation.

## 1 INTRODUCTION

In the last decades industrial robots have already demonstrated their usefulness in many sectors of production. Their deployment made fast, vast, and cheap production possible. However, due to safety reasons close/direct cooperation is traditionally not considered. In order to increase competitiveness in production the human and the robot should work as a team to achieve more flexible human-robot interaction [1]. But how do operators experience this close cooperation with a robot in the assembly line? According to the definition of Alben [2], User Experience (UX) comes everywhere into play where humans interact with a system. This broadens the scope of UX to concepts such as cooperation and usability, including factors such as perceived safety, stress, or emotion. Especially in the factory context UX aspects are crucial as they might affect the production process [3]. Research in HRI on factory robots showed that UX factors are not static, but change over time [5].

The case study presented in this paper builds on this previous research by investigating the UX of a cooperative robot that was deployed for three weeks in the assembly line of a factory. We used the established phenomenologically oriented interview guidelines from [5] in order to gather the impressions, emotions, and thoughts of the workers with the aim of improving the human-robot collaboration experience for the future. This qualitative approach especially focusses on the characterization of workers' experiences, taking into account expectations and familiarization issues.

[1] All authors are with the Vienna University of Technology, Austria.
Email: {firstname.lastname}@tuwien.ac.at

## 2 RELATED WORK

The Industry 4.0 paradigm envisions a "Smart Factory" in which humans and robots will work more and more closely together. Cobots are robots designed for direct physical interaction with the workers while their design is research subject until today [9]. Cobots work in close cooperation with the users and can overcome inertia and friction forces, but do not allow pre-programmed (autonomous) movements. Therefore they have safely limited speeds or forces, automatic restart and allow to guide the robot by hand. Robot systems that can be operated without spatial safety areas gain acceptance in the market since 2010 [10][11]. Bogh et al. [12] present a historical overview of autonomous robot-based manipulation systems and show its intense research for the last 30 years. The sparse industrial use of autonomous manipulation systems is based on the conservative attitude of the industry (especially with regard to safety and standardization issues), the high system costs, the lack of flexibility as well as on the low feasibility of basic research results (technology transfer gap - especially in Europe [13]).

Therefore, it is necessary to research how workers experience the interaction with robots. We are not aware of much experience research in the context of industrial robotics. A quantitative questionnaire study on UX in the area of industrial robots [3] already showed that the UX of robots without safety fences changes over time, from a rather negative experience towards a more positive one, which however, never reaches the same level of positive experience as robots in safety fences with which the workers do not have to collaborate. A qualitative follow-up study [5] revealed more insights on the transition effects of the workers' experiences regarding the expectations before deployment, the familiarization with the robots, and the experienced consequences of working with the robots. Other studies on the expectations towards industrial robots showed that the fear of being replaced has a crucial impact on how a robot is accepted and experienced in the industrial context [4],[6].

The focus of our research followed a similar interest, namely the UX of a newly introduced robot without a safety fence in the assembly line of an automotive factory. The aim of our case study was two-fold (1) collecting suggestions for improvement of the human-robot collaboration for the next prototype iteration and (2) adding a building block to the existing knowledge on the dynamics of UX in human-robot cooperation in the factory context.

The research presented in this paper is part of a 2-years research project with the goal to develop an assistive robot system, which is characterized by its feasibility for Industry 4.0 environments in two different testbeds (automotive assembly and polishing of casting molds). In a multistage process, the integration of end users in the development process should contribute to the definition of interaction paradigms going beyond existing solutions. The first iteration (stage) is based on a robot system off-the-shelf from Universal Robots in order to evaluate the industrial feasibility of the state of the art. In this context the interviews presented in this paper took place. The system of the second iteration will be derived from the results of user studies in both testbeds, involving additional sensors for object localisation and path correction. The third development iteration will represent the full expansion stage and will integrate the recommendations of all previously conducted user-centered studies in order to obtain a robot-based assistive system that is feasible for Industry 4.0 requirements. Our overall approach is strongly influenced by [14] and prioritized the participation of all partners in all phases of usability evaluation.

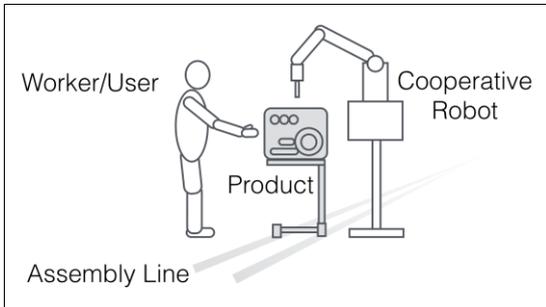

**Figure 1.** Schematic illustration of the human-robot collaboration set-up during the assembling process on the assembly line.

Collaborating with the robot in our use case basically meant that a robot (a collaborative robot from the current Universal Robots series) tightens screws, which the worker firstly had to position. The worker puts a motor block on the working space, puts the screws in the right place and then the robot tightens them and the worker checks (and if necessary improves) the quality afterwards. This is one station in the overall assembly line (see Figure 1) and during one shift workers switch regularly between stations (whereas this is the only one with this kind of human-robot collaboration involved). Before the robot was introduced in this work station, the worker did all the steps (putting the motor block and screws in place and tighten them) manually. The robot was introduced to the participants as a research prototype that will be explored over a period of three weeks for testing its general suitability in the assembly line and that it will be improved in terms of usability over the next 1.5 years before it will be most likely permanently installed in this work station.

## 3 METHODOLOGY AND PROCEDURE

Methodologically we based our interviews on the guidelines presented in [5]. These interview guidelines are intended to cover situational, descriptive, evaluative, and argumentative information in order to get a proper image of the context, the routines, and subjective perceptions and explanations. The guideline structures the interview in order to get insights into (a) general aspects regarding the work with the robot, (b) experiences before the introduction of robots, (c) first confrontation with the new system/enrollment, (d) special experiences working with the new systems, and (e) special aspects of the HRI (e.g. safety).

We chose this methodology, as narrative interviews offer an open structure which can evoke personal experiences [7]. The interviewer stimulates interviewees to report on these by asking so-called trigger questions. During the participant's report the interviewer should mainly be a listener and not influence the story. Additionally to trigger questions the interview guidelines also involved so-called semantic questions [7], which should not trigger experience reports, but argumentations regarding the interviewees' ascriptions of meaning. An example for a semantic question is "What does 'robot' mean to you?".

Interviewees were workers who had been collaborating with the robotic prototype for three weeks in the assembly line. They were directly affected by the transition from working without robots to working with them. Our industrial partner recruited the five participants for us. The procedure was as follows: The interviewer introduced the participants to the study purpose and goal. Each participant filled in an informed consent form. After the main interview participants filled in a short demographic questionnaire and were thanked for their participation.

The interviews were recorded and analyzed by the authors of this paper, who are independent of the robot developers and the factory administration, following a thematic analysis approach [8]. The aim of the analysis was the structuring and interpretation of the data and the derivation of findings. We used no predefined categories, but the relevant subjective meanings and experiences were identified through an exploratory and data-focused analysis. After a rough identification of relevant issues, a more detailed description of each topic was made. In order to derive final categories the different topics of the participants were compared. In a second step overlapping topics were merged to categories or clearly differentiated from each other.

## 4 RESULTS

A total of five assembly workers were recruited for the study, who beforehand participated in a user study in which they learnt how to teach the robot waypoints for tightening the screws. For details on the participants' characteristics see Table 1.

| No. | Gender | Age | Work experience with robot in hours |
|---|---|---|---|
| 1 | Female | 27 | 21 |
| 2 | Male | 61 | 13 |
| 3 | Male | 40 | 15 |
| 4 | Male | 48 | 8 |
| 5 | Female | 50 | 12 |

**Table 1.** Participant details.

All participants were experienced with computers and automated systems previously to collaborating with the robot. All except one participant were right handed and assessed their spatial imagery as average or above. Each participant was interviewed in German (mother tongue of all participants) for 30 minutes. During the analysis of the recorded interviews the expressed impressions, emotions, and thoughts of the participants were clustered into several main categories. Table 2 shows the most prominent clusters which have emerged from this process.

| No. | Categories | Stated by |
|---|---|---|
| 1 | The robot determines my work and speed. | 4 out of 5 |
| 2 | The robot should be more adaptive. | 4 out of 5 |
| 3 | The robot does not stop at physical contact. | 4 out of 5 |
| 4 | To work with the robot was a big transition. | 3 out of 5 |
| 5 | The robot is a helpful tool. | 2 out of 5 |

**Table 2.** Most prominent topic clusters on the human-robot collaboration experience.

The first two findings "the robot determines my work and speed" and "the robot should be more adaptive" represents findings with respect to experienced consequences of working with the robot. Four out of five participants negatively experienced it that the robot changed their personal way of doing the task of screw-tightening. Due to the robot there was a specific speed and rhythm in which the task had to be performed. *"The robot slows me down… especially during the assembling group D where I have to wait until it has finished its working sequence."* (Participant 2). This finding has to be interpreted in the context of the overall working procedure of the assembly line, which is highly dynamic and can individually be adapted to the preferences and ergonomic needs of the person. An aspect which got reduced through the robot.

Similarly, experienced changes in the work routines were observed in [5]. However, in this study the context was a semi-conductor factory and the introduction of the robot actually changed the sequence of actions, which was negatively experienced as an increase in complexity. Nevertheless, in both studies participants wished for more adaptation of the robot, therefore the workers do not have to adapt themselves.

Interestingly, this change in the working routine was also perceived as a shift in their social life. *"Usually, there is some time in-between for some chit-chat with my colleagues. However, this is nearly impossible while I am working with the robot. It completely determines my working rhythm."* (Participant 4). Before the deployment of the robot, the dynamics of the assembly line allowed that at every station people worked in their own rhythm and timing and thereby they could use short pauses for chatting with colleagues. Losing that option was experienced as a downside of the robot station. However, two out of the five participants have already acknowledged just after three weeks and in the first prototypical installation the added value of the robot as a tool in the assembly line. Still the physical and mental requirements during the introduction of the robot were stated as high, but a quick learning curve was experienced as well.

Finally, the rigidity of the system led to a feeling of mistrust in its safety (4 out of 5). However, low ratings of perceived safety for newly introduced robots without safety fences were also found in [3], but it turned out that over time and with more experience safety ratings increased.

Overall, as also reported in [5] the interviews were appreciated by the workers as an opportunity for expressing their thoughts on the system and for involving them in the further development of the next prototype stage. They valued it as being seen as experts for the working routines in which the robot should be integrated.

## 5. CONCLUSION & FUTURE WORK

The presented case study demonstrated that the given context of the future robot users strongly determined the experience of the robot's capability. This was shown by investigating the UX of a newly introduced robotic prototype after three weeks of its deployment by the means of narrative interviews. We found that the newly introduced robotic system was not well-perceived with regard to its adaptivitiy towards the users' individual rhythm, speed and working steps. It can be assumed that this essential lack in flexibility is at least partly responsible for the other mentioned shortcomings in perceived safety, usability, and general helpfulness.

Basically, the introduced cooperative robot in its current prototype stage bears the risk to re-establish a rigid production line logic, which does not ideally foster the smart factory paradigm. Our results support the findings of others [5] that cooperative robots in a dynamic factory context have to adapt to their human co-workers by taking their individual working steps and speed into account. This can be realized by additional sensors for posture recognition in order to measure the worker's movements and waiting positions, which may improve the factor of perceived safety as well. This approach would help easing the transition of working with a new robot and increase its overall UX. Additionally, further research on UX over time is planned after a technical revision of the system in order to get a deeper understanding of the cooperation, perceived safety, perceived usability, stress, general UX, and emotion.

## ACKNOWLEDGEMENTS

This work was funded by "Produktion der Zukunft" a programme of the Federal Ministry of Science and Research of Austria (848653 - AssistMe). We especially want to thank our colleagues from Profactor, Jürgen Minichberger and Markus Ikeda for the robot integration in the factory assembly line.

## REFERENCES

[1] Weiss, A., Buchner, R., Fischer, H., and Tscheligi, M. Exploring human-robot cooperation possibilities for semiconductor manufacturing. In Int. Workshop on Collaborative Robots and Human Robot Interaction (2011)

[2] Alben, L.: Quality of experience: defining the criteria for effective interaction design. Interactions 3 (3), 11–15 (1996)

[3] Buchner, R., Wurhofer, D., Weiss, A., & Tscheligi, M. (2013). Robots in time: How user experience in human-robot interaction changes over time. In Proceedings of ICSR2013, pp. 138-147.

[4] Obrist, M., Reitberger, W., Wurhofer, D., Förster, F., Tscheligi, M.: User experience research in the semiconductor factory: a contradiction? In Proceedings of INTERACT 2011, pp. 144–151 (2011).

[5] Wurhofer, D., Meneweger, T., Fuchsberger, V. and Tscheligi, M. Deploying Robots in a Production Environment: A Study on Temporal Transitions of Workers' Experiences. In Proceedings of INTERACT 2015, pp. 203-220 (2015).

[6] Weiss, A., Igelsböck, J., Wurhofer, D., Tscheligi, M.: Looking forward to a "Robotic Society"? International Journal of Social Robotics, 3(2), 111-123. (2011).

[7] Flick, U.: Episodic interviewing. In: Bauer, M.W., Gaskell, G. (eds.) Qualitative Researching with Text, Image and Sound, pp. 75–92. Sage, London (2000) USA.

[8] Braun, V., Clarke, V.: Using thematic analysis in psychology. Qual. Res. Psychol. 3 (2), 77– 101 (2006)

[9] Gambao, E.; Hernando, M.; Surdilovic, D. A new generation of collaborative robots for material handling. Gerontechnology, 11(2), pp. 368 (2012).

[10] http://www.mrk-systeme.de/index.html

[11] http://www.kuka-labs.com/de/service_robotics/lightweight_robotics/

[12] Bøgh, S., Hvilshøj, M., Kristiansen, M., & Madsen, O. (2011). Autonomous industrial mobile manipulation (AIMM): from research to industry. In 42nd International Symposium on Robotics.

[13] Griffiths, S., Voss, L., & Rohrbein, F. (2014, May). Industry-Academia Collaborations in Robotics: Comparing Asia, Europe and North-America. In Robotics and Automation (ICRA), 2014 IEEE International Conference on (pp. 748-753).

[14] Buchner, R., Mirnig, N., Weiss, A., & Tscheligi, M. (2012, September). Evaluating in real life robotic environment: Bringing together research and practice. In *RO-MAN, 2012 IEEE* (pp. 602-607).